# Diagnostic Accuracy and Navigation Behavior of Vision-Language Model Agents Reading Emergency CT and MRI Examinations

Jonathan Lichtle[1*], Aymen Meddeb[2*], Antonin Birot[1], Ibrahima Niang[3], Lisa Gastaud[1], Riyad Hanafi[1], Sofiane Dujardin[1], Corentin Selomme[4], Guillaume Cachard[4], Aghiles Hamroun[5,6], Grégory Kuchcinski[1,7], Su Hwan Kim[8], Bastien Le Guellec[1,6,7†]

[1] Neuroradiology department, Lille University Hospital, Lille, France
[2] Neuroradiology department, Charite, Berlin, Germany
[3] Radiology department, Cheikh Anta Diop University Hospital, Dakar, Senegal
[4] Abdominal radiology department, Lille University Hospital, Lille, France
[5] Public health department, Lille University Hospital, Lille, France
[6] UMR1167 RID-AGE, Pasteur Institute of Lille, Inserm, Lille University, Lille University Hospital Center, Lille, France
[7] INSERM, U1172-LilNCog-Lille Neuroscience & Cognition, Université de Lille, Lille, France
[8] Radiology department, Technical University Munich Hospital, Munich, Germany

* Equal contribution. † Corresponding author: bastien.leguellec@chu-lille.fr

## Abstract

**Purpose.** To evaluate the diagnostic accuracy and image navigation behavior of vision-language model (VLM) agents under varying exploration strategies.

**Materials and Methods.** In this retrospective study, six VLMs read 100 emergency CT and MRI examinations published in a public archive (March 2025–August 2026) under four conditions (free navigation, navigation from a key slice, and all slices or key slices in a single request), twice each (4800 runs). Nine radiologists read 30 of those cases. A large language model judge validated against two radiologists graded the answers. Conditions were compared with generalized estimating equations clustered by case.

**Results.** Pooled accuracy was lowest under free navigation (32.6%; 391/1200), followed by navigation from a key slice (37.9%; 455/1200; $P = .003$), and highest with all slices (42.2%; 506/1200) or key slices (42.5%; 510/1200) in one request (both $P < .001$). On the 30 cases read by radiologists, freely navigating models were correct in 30.0% of runs (108/360) versus 77.0% for radiologists (208/270; $P < .001$). Models formed a first hypothesis after a median of 10.8% of the slices and kept it in 63.1% of runs (755/1196); final answers were correct in 12.4% of runs after a wrong first hypothesis (110/885) versus 90.4% after a right one (281/311; $P < .001$). Median stated confidence in the final answers was 90%, whereas 38.9% of them were correct (1861/4778).

**Conclusion.** Across all exploration strategies, VLM agents remained far less accurate than radiologists. They adopted an early hypothesis and usually kept it, seldom correcting a wrong one.

## Introduction

Vision-language models (VLMs) capable of analyzing medical images have caught the attention of radiologists since the release of GPT-4V in 2023, although their diagnostic accuracy remains below that of radiologists (1–3). However, these models are still evaluated almost exclusively on preselected key images selected by radiologists, which bypasses the information gathering of real-world interpretation. Radiologists explore volumetric datasets, decide where to look, switch between planes, and integrate new findings to refine their differential diagnosis through search, recognition, revision and decision (4–6).

This search can also introduce errors: cognitive biases account for 74% of image interpretation errors (7). Some arise from the exploration itself, including premature closure, in which a reader commits to a diagnosis before the discriminating information has been examined, and anchoring, in which a reader fails to revise an early hypothesis despite contradictory findings (7,8). These failures are well studied in radiologists, but not in autonomous AI systems placed in a similar environment.

Agents, models capable of interacting with data with tools, have recently been developed to perform autonomous navigation of radiology examinations (9–12). Their safe deployment requires evaluating the failure modes specific to this way of reading images. In text-based clinical tasks, large language models diagnose less accurately when they must gather information themselves than when it is provided upfront, and their accuracy depends on the order in which this information arrives (13) and drops when questions are worded to induce cognitive biases (14). Whether VLM agents exploring an imaging examination are similarly affected, and whether their exploration shows behavioral analogs of premature closure and anchoring, the cognitive errors described in radiologists, is unknown.

This work aimed to evaluate the diagnostic accuracy and image navigation behavior of VLM agents under varying exploration strategies.

## Materials and Methods

This retrospective study is reported according to the Checklist for Artificial Intelligence in Medical Imaging (CLAIM) (15). Because it used only publicly available, fully anonymized data from Radiopaedia, the study was exempt from institutional review board approval.

### Case Selection

We searched cases published on Radiopaedia (radiopaedia.org) between March 2025 and August 2026, filtered by system and modality (CT and MRI) without keywords (Appendix S1). A radiology resident and a radiologist (A.B. and J.L.; 3 and 4 years of experience) selected studies that could be encountered in the emergency department, and classified them in four anatomical domains (abdominal, neurologic, thoracic, and musculoskeletal and trauma) (16–18). Cases could include series from another examination when the diagnosis required them (Appendix S1). They were presented to models and radiologists with a one-line clinical indication written from the case history (Table S1). Readers and models never saw the case title, diagnosis or discussion. The reference standard was the diagnosis published with the case (diagnostic certainty: certain, 65 cases; almost certain, 30; probable, 5). They also marked key slices, including one most informative slice, which a third radiologist (B.L.G.; 7 years of experience) reviewed.

### Radiologist Readers

Nine radiologists from four institutions (three senior residents and six board-certified radiologists; median experience, 7 years [range, 3–10]; subspecialties in Appendix S1) independently read 30 of the cases in a web viewer that showed the same stacks, window and indication, allowed the same one-slice moves and series switches as the agents, and logged every action. Reading behavior was analyzed up to the first saved diagnosis. Readers were blinded to the source of the cases and took no part in selecting the cases they read, marking their key slices, or grading. Their answers were graded by the same large language model judge as the models' answers and, as a sensitivity analysis, by two radiologists blinded to the reader, who resolved disagreements by consensus.

**Models**

We selected six VLMs from six developers, according to their rank on the LLM Arena leaderboard (https://lmarena.ai/leaderboard) on August 25, 2026, the date of the first run, among models priced below US $2 per million output tokens (Appendix S1): Gemini 3.7 Flash (Google), GPT-5.6 Luna (OpenAI), GLM-5.3 Flash (Z.ai), Qwen 3.7 Flash (Alibaba), MiniMax M3 (MiniMax) and MiMo v2.5 (Xiaomi). All were accessed through the OpenRouter API in August–September 2026, at temperature 0 with the provider's default reasoning setting. As a safety check, accuracy was compared between cases published before and after each published knowledge cutoff when it was available.

**Agent Environment**

The agent environment reproduced the radiologists' viewer (Fig 1). At each turn, the agent received the current slice, its position and the result of its last command, and replied with five fields: check (what the slice shows), adjust (how this changes its understanding), plan (what to do next), a working hypothesis with a confidence from 0 to 100, and a command: move one slice up or down, switch series, or answer. This process is derived from the PDCA cycle (19). Only the three most recent slices remained in the models’ memory. As a sensitivity analysis, free navigation on the 30 cases read by radiologists was repeated with every image kept in context. The hidden reasoning tokens were not accessed. Runs were fully autonomous. Most navigation runs (2103/2400) answered before a step limit of 1.2 times the number of slices; the other 297, mostly by MiniMax M3 and Qwen 3.7 Flash, were then asked for an answer, and letting them continue up to three times the number of slices did not change accuracy (Appendix S2). The final answer gave a normal or abnormal call, a primary diagnosis (“Normal study” was allowed), a differential diagnosis and a confidence. Only the primary diagnosis was graded. Prompts and parameters are given in Appendix S2.

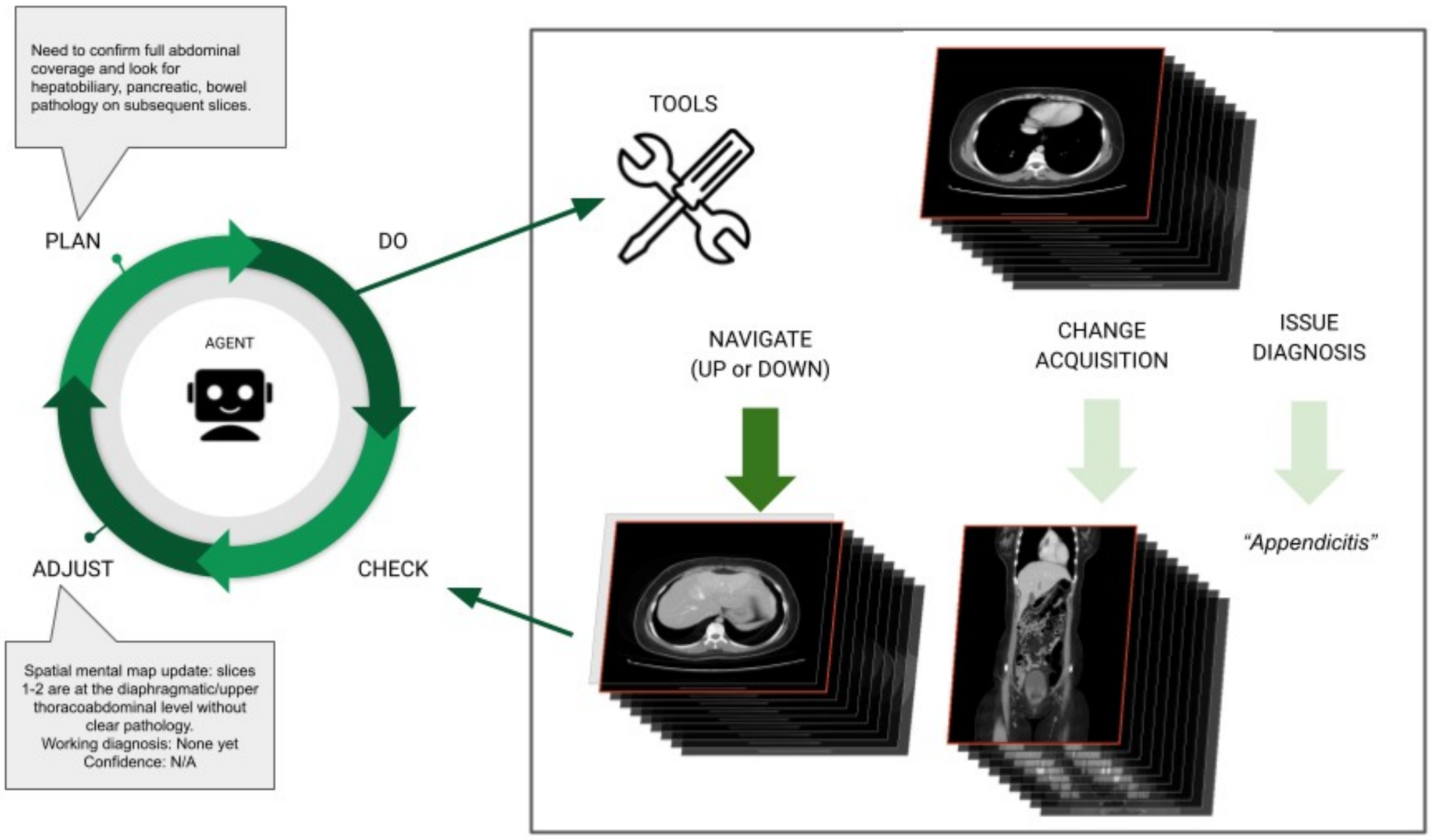

**Figure 1:** Agent environment. At each turn, the agent describes the current slice (check), updates its understanding of the case (adjust), states its next step (plan), gives its working hypothesis with a confidence, and issues one command (do): move one slice, switch series or answer.

### Reading Conditions

In free navigation, the agent started from a series of its choice. In navigation from the key slice, it started on the most informative key slice. In the two single-request conditions, all slices (median, 70 images per case) or only the key slices were sent at once, labeled by series and slice number. As a no-image control, each model also read every case once with the slice labels but no images, and with the indication alone.

### Grading of Answers

A large language model judge, Claude Sonnet 5 (Anthropic), from a developer whose models were not evaluated, graded each primary diagnosis as correct, partial or incorrect against the reference, blinded to model and condition. On 500 runs graded independently by two radiologists (J.L. and B.L.G.; disagreements resolved by consensus), it agreed with the consensus as closely as each radiologist did ($\kappa$ = 0.83, vs 0.84 and 0.83) and better than three other candidate judges (Appendix S1). The judge also classified every wrong answer by the location of the named abnormality relative to the reference (Appendix S2). Two radiologists blinded to model and condition checked 100 of these labels and both confirmed all 100.

### Behavioral Measures

For each navigation run and radiologist read, we measured the share of slices viewed, the number of navigation actions, actions per slice viewed (1 when no slice is revisited), whether a key slice was displayed, and a dwell ratio on key slices (1 for a uniform sweep; Appendix S1). Claude Sonnet 5 divided each run's sequence of working hypotheses into successive distinct diagnoses, identifying the first hypothesis and whether the final answer was the first. On 100 runs segmented independently by two radiologists blinded to this output (B.L.G. and A.M.), agreement with it on whether the first hypothesis was kept was κ = 0.67 and 0.78, similar to the agreement between the two radiologists (κ = 0.63; Appendix S1).

### Statistical Analysis

The unit of analysis was the run, and runs of the same case were treated as correlated. Runs without a usable answer (18/4800; provider content filter) were counted as incorrect. Conditions were compared with logistic generalized estimating equations (GEE) clustered by case and adjusted for model (omnibus Wald test, then Holm-adjusted pairwise odds ratios). Other comparisons used logistic or Gaussian GEE. Ninety-five percent CIs came from bootstrap resampling of cases (10 000 resamples), or of cases and readers for comparisons with radiologists. After their first mention, 95% CIs are given as unlabeled lower–upper bounds. Agreement was measured with Fleiss κ, and calibration with the Brier score and the area under the receiver operating characteristic curve (AUROC). The comparison of conditions was the primary analysis; other analyses were exploratory. A two-sided $P < .05$ indicated statistical significance. Analyses used Python 3.12 (statsmodels 0.14.5, SciPy 1.16.1; Appendix S3).

## Results

The 100 cases (Table S1) comprised 31 abdominal, 30 neurologic, 20 thoracic, and 19 musculoskeletal and trauma examinations (73 CT, 26 MRI, and 1 CT with diffusion-weighted MRI), seven of them normal studies.

### Diagnostic Accuracy

Navigation condition significantly altered accuracy (Wald $P < .001$; Table 1, Fig 2a). Pooled across the six models, accuracy was lowest under free navigation (32.6% [391/1200 runs]; 95% CI: 26.7%, 38.8%) and increased when navigation started on the key slice (37.9% [455/1200]; odds ratio [OR], 1.27 [1.11–1.46]; $P = .003$) and when all slices (42.2% [506/1200]; OR, 1.52 [1.34–1.73]) or only the key slices (42.5% [510/1200]; OR, 1.54 [1.29–1.85]) were sent in a single request (both $P < .001$). Navigation from the key slice remained less accurate than both single-request conditions (Holm $P = .01$ and .03), which did not differ from each other ($P = .86$). Free navigation was the least accurate condition for four of the six models and tied for a fifth (Fig 2c; condition-by-model interaction, $P = .02$) and in all four anatomical domains, with a similar penalty in each ($P = .93$), although accuracy varied between domains, from 30.9% (282/912 runs) for musculoskeletal and trauma to 53.1% (510/960) for thoracic cases ($P = .05$; Table S2). On the seven normal studies, 58.9% of runs (198/336) were correct. On the 34 cases published after the latest knowledge cutoff, free navigation remained least accurate (30.4% [124/408] vs 41.9% [171/408] with all slices; OR, 1.67 [1.36–2.05]), and accuracy did not differ between cases published before and after each published cutoff ($P = .15$, .88 and .92). Agreement between repeated runs was lowest under free navigation for four of the six models (Fleiss κ, 0.32–0.85; Table S3).

**Table 1: Diagnostic Accuracy by Reading Condition**

| Reading condition | Correct, n/N | Accuracy, % (95% CI) | Comparison (95% CI) | P value |
|---|---|---|---|---|
| **All 100 cases, six models (two runs per case)** | | | | |
| Free navigation | 391/1200 | 32.6 (26.7–38.8) | Reference | — |
| Navigation from the key slice | 455/1200 | 37.9 (31.8–44.4) | OR, 1.27 (1.11–1.46) | .003 |
| All slices, single request | 506/1200 | 42.2 (35.7–48.8) | OR, 1.52 (1.34–1.73) | <.001 |
| Key slices, single request | 510/1200 | 42.5 (36.2–48.9) | OR, 1.54 (1.29–1.85) | <.001 |
| **No-image controls, all 100 cases (one run per case)** | | | | |
| Slice labels, no images | 103/600 | 17.2 (12.0–22.5) | OR, 0.28 (0.20–0.38)* | <.001 |
| Clinical indication only | 68/600 | 11.3 (7.2–15.7) | OR, 0.17 (0.11–0.25)* | <.001 |
| **30 cases read by the radiologists (models, two runs per case; nine radiologists)** | | | | |
| Radiologists | 208/270 | 77.0 (65.9–87.0) | Reference | — |
| Radiologists, graded by two radiologists | 223/270 | 82.6 (71.1–91.9) | — | — |
| Free navigation | 108/360 | 30.0 (19.2–41.1) | −47.0 points (−60.2 to −33.9) | <.001 |
| Navigation from the key slice | 124/360 | 34.4 (22.2–46.9) | −42.6 points (−55.7 to −29.7) | <.001 |
| All slices, single request | 145/360 | 40.3 (29.2–51.9) | −36.8 points (−48.8 to −24.9) | <.001 |
| Key slices, single request | 163/360 | 45.3 (33.6–57.2) | −31.8 points (−44.3 to −18.8) | <.001 |

Note.—Accuracy is the share of runs (models) or reads (radiologists) whose primary diagnosis was correct; 95% CIs from a bootstrap resampling cases (and readers for radiologists). Odds ratios (ORs) are relative to free navigation (*relative to all slices in a single request, the same request with images), from logistic generalized estimating equations clustered by case and adjusted for model; P values for the main conditions are Holm adjusted. Model runs and radiologists' reads were graded by the same model judge; the second radiologist row gives the grading by two radiologists. Differences from radiologists are in percentage points, with P values from the bootstrap. Runs without a usable answer were counted as incorrect.

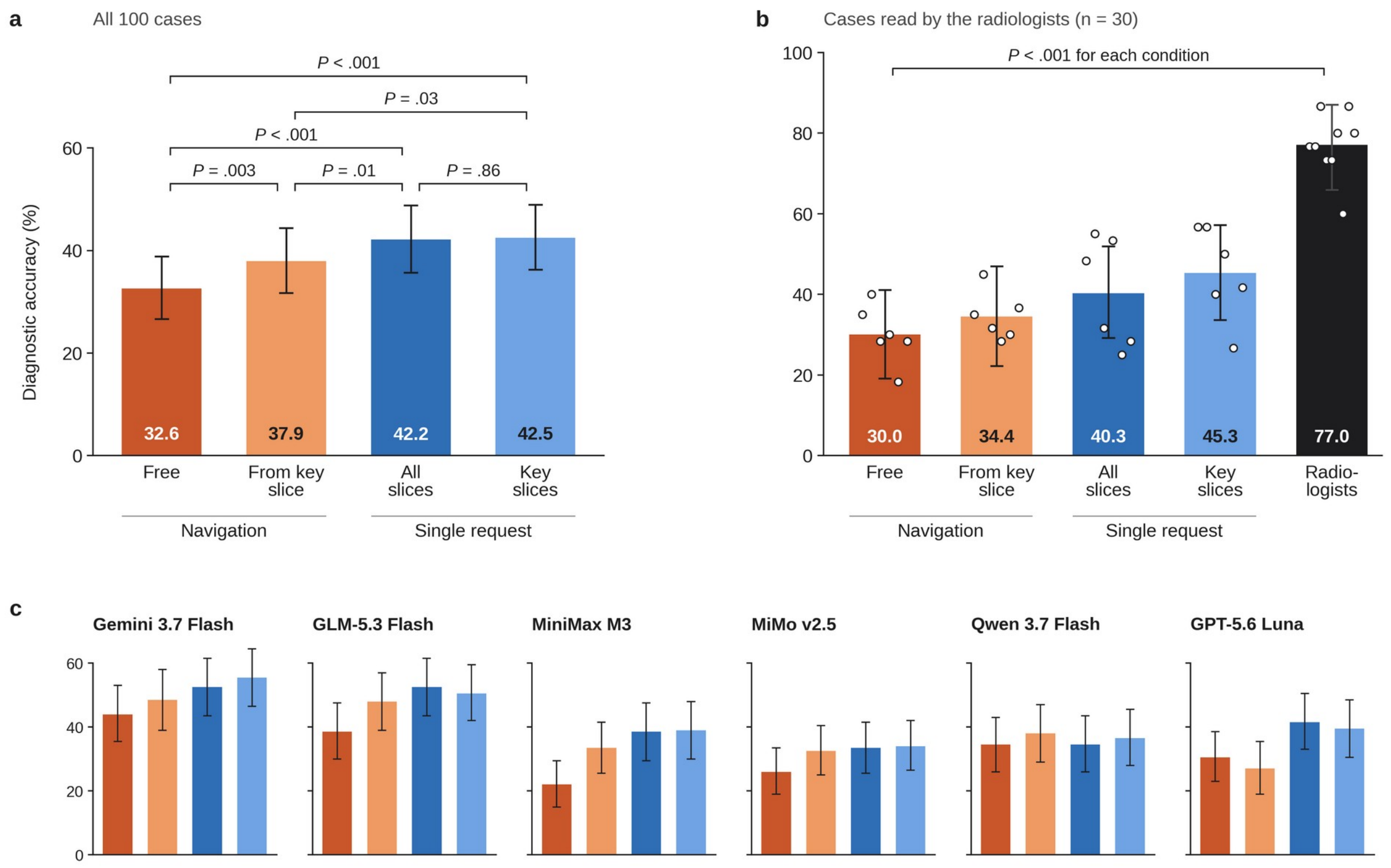

**Figure 2:** Diagnostic accuracy under four reading conditions, and comparison with radiologists. (a) Pooled accuracy of the six models on all 100 cases (1200 runs per condition); brackets give Holm-adjusted P values (logistic generalized estimating equations clustered by case, adjusted for model). (b) The 30 cases read by the radiologists: pooled model accuracy under each condition (360 runs per condition) and the nine radiologists (270 reads, free navigation), all graded by the same judge; dots, individual models and radiologists; P values from a bootstrap resampling cases and readers. (c) Accuracy of each model on all 100 cases (200 runs per condition); the four models on the left had their lowest accuracy under free navigation. Free and From key slice: navigation from a blank start or from the key slice; All slices and Key slices: one request. Numbers in the bars are accuracies (%). Error bars, 95% CIs (bootstrap resampling cases, and readers for the radiologists).

## No-Image Control

Without any image, models were still correct in 17.2% of runs given the slice labels (103/600; 12.0%–22.5%) and in 11.3% given the indication alone (68/600; 7.2%–15.7%), compared with 42.2% with all slices (both $P < .001$; Table 1). Only 5.5% of these answers (33/600) reported that the images were missing. On the 54 cases that no model answered correctly without images, freely navigating models were correct in 18.2% of runs (118/648), compared with 49.5% (273/552) on the other cases (Table S4).

## Error Analysis

About half of the wrong answers that named an abnormality placed it in the correct organ (54.5% [1161/2132] with images), with no difference between free navigation and single-request reading ($P = .25$). Without images, this share fell to 39.9% (114/286; OR, 0.52 [0.38–0.72]; $P < .001$; Table S5).

## Comparison with Radiologists

On the 30 cases read by radiologists, freely navigating models were correct in 30.0% of runs (108/360; 19.2%–41.1%) and the nine radiologists, graded by the same judge, in 77.0% of reads (208/270; 65.9%– 87.0%; *P* < .001; Fig 2b). The other conditions were also less accurate than the radiologists (differences, 31.8 to 42.6 percentage points; all *P* < .001; Table 1). Graded by two radiologists, the radiologists were correct in 82.6% of reads (223/270). Agreement among radiologists was fair (Fleiss κ = 0.22; pairwise agreement, 72.4%).

### Exploration Patterns

The share of the examination viewed before answering differed between models (*P* < .001; Fig 3a). On the 30 cases they read, radiologists viewed a median of 77.9% of the slices (IQR, 63.5%–90.9%). On the same cases, Gemini 3.7 Flash stopped earlier, viewing 43.5% (IQR, 30.0%–67.4%; difference, 34.4 percentage points [22.9–43.1]; *P* < .001), whereas the five other models viewed 100% (IQR, 77.1%–100%; difference, 22.1 percentage points [15.0–29.9]; *P* < .001), in a single pass through each series (1.00 action per slice viewed vs 1.81 for the radiologists; Fig 3b). Results were similar on all 100 cases (Fig S1).

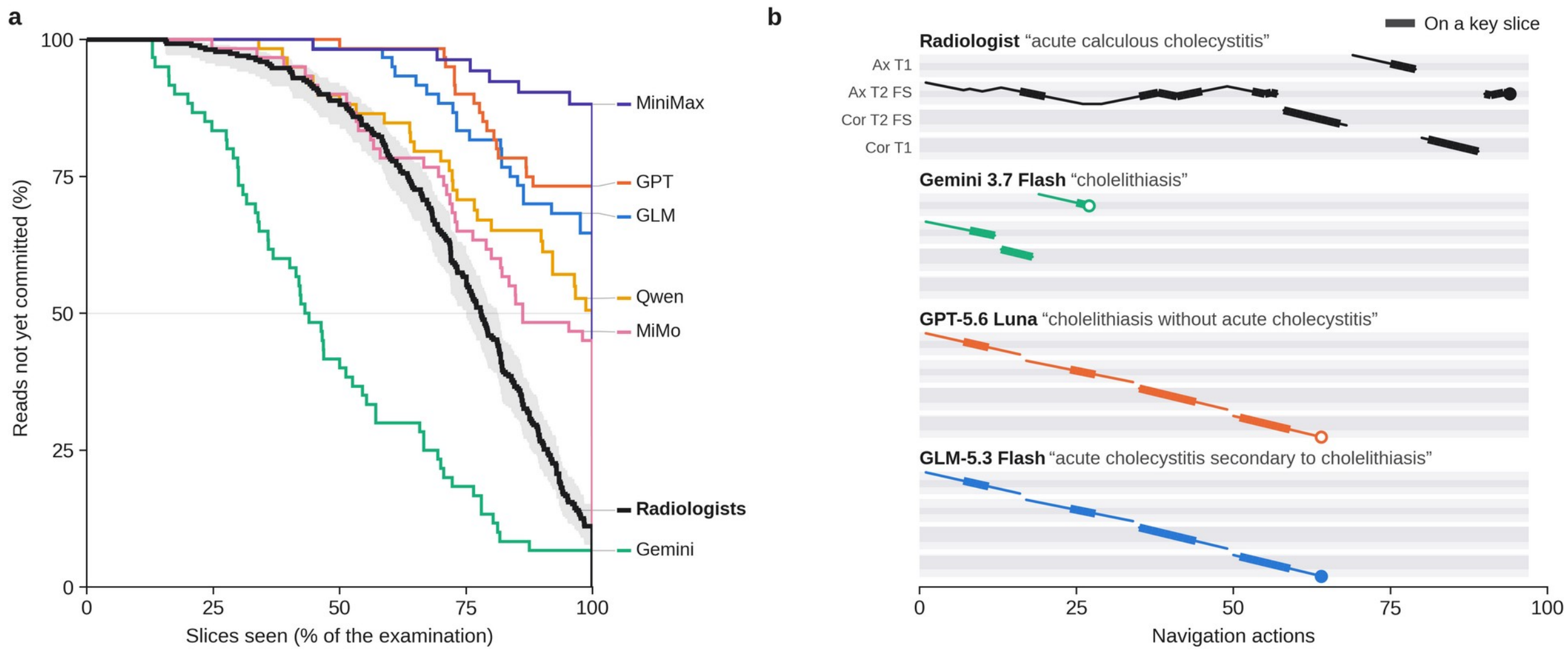

**Figure 3:** Share of the examination seen before answering, and one examination read by a radiologist and three models (free navigation, the 30 cases read by the radiologists). (a) Kaplan-Meier curves of reads not yet committed to a diagnosis against the share of slices seen. Black, nine radiologists (270 reads; first saved diagnosis; grey band, 95% CI); colored, each model (60 runs). (b) The example examination (acute cholecystitis, four MRI series), one row per series: the line gives the slice on screen after each action and is thicker on key slices (faint grey bands); the end dot is filled for a correct answer and open for an incorrect one. Ax = axial, Cor = coronal, FS = fat saturated.

Exploration was rarely the limiting step: 95.7% of free-navigation runs (1144/1196; 93.4%–97.6%) displayed a key slice before answering, including 95.3% of runs with a wrong answer (767/805). On shared cases, models and radiologists displayed a key slice equally often (98.1% [353/360] vs 99.3% [268/270]; difference, 1.2 percentage points [−1.4 to 3.3]). Radiologists spent more of their actions on key slices than a uniform sweep would (median dwell ratio, 1.27 [IQR, 1.04–1.54]), whereas models did not (1.00 [IQR, 0.96–1.09]; *P* < .001).

### First Hypothesis

Although the five models other than Gemini 3.7 Flash answered after viewing a median of 98.1% of the slices on all 100 cases, their first hypothesis emerged after a median of 10.8% of the slices (9.1%–13.3%; range across models, 7.8%–16.7%; Fig 4a). It was correct in 26.0% of runs (311/1196; 20.7%–31.6%) and was kept as the final answer in 63.1% (755/1196; 59.3%–66.9%; Fig 4b). The final answer was correct in 90.4% of runs after a right first hypothesis (281/311; 86.0%–94.1%) but in 12.4% after a wrong one (110/885; 9.1%–16.1%; OR, 29.0 [17.6–47.7]; *P* < .001). A wrong first hypothesis was kept in 55.8% of runs (494/885) and a right one in 83.9% (261/311); of the 391 runs that left a wrong first hypothesis, 110 reached the correct diagnosis.

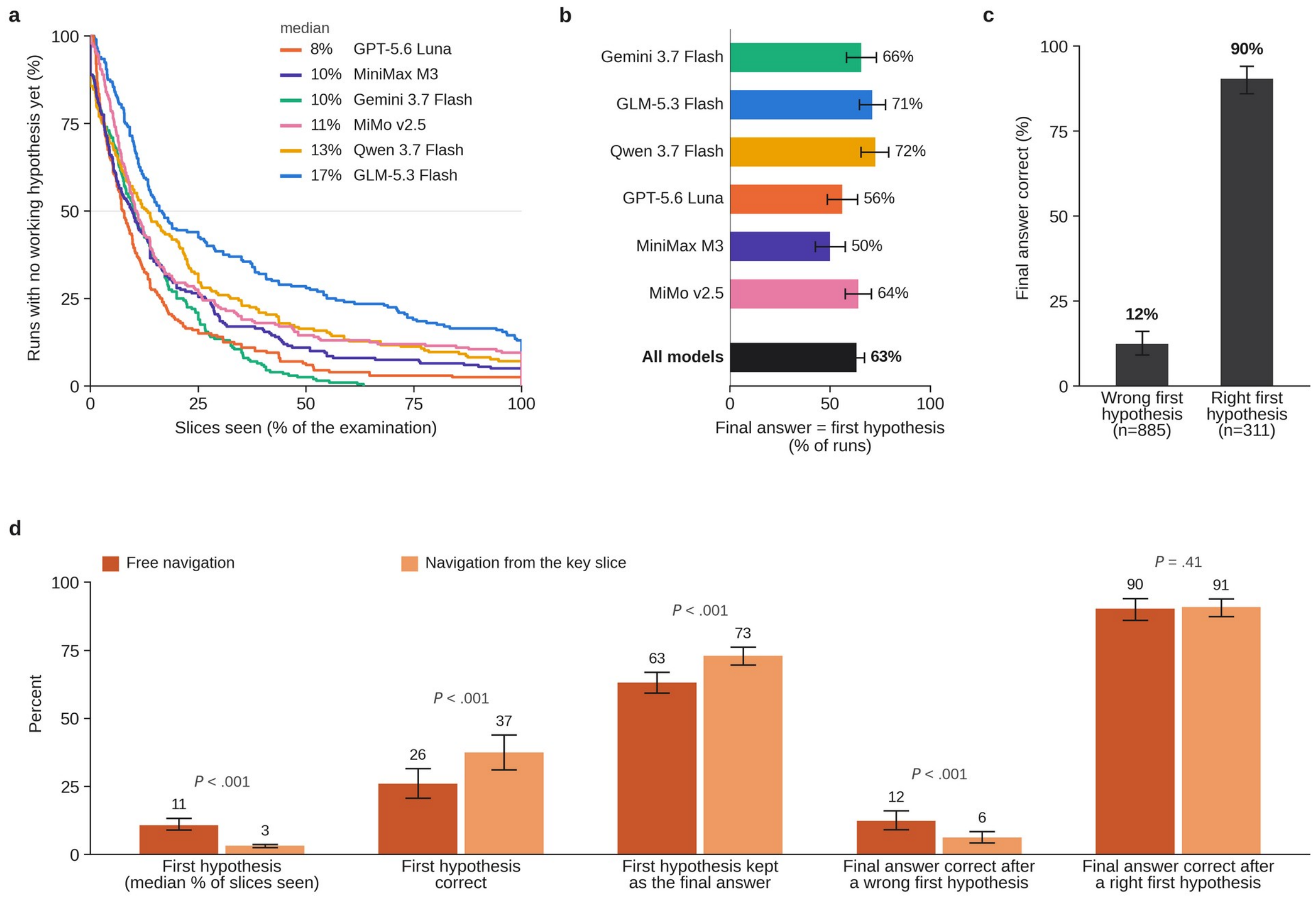

**Figure 4:** First working hypotheses during free navigation (six models, 100 cases, 1196 runs). (a) Kaplan-Meier curves of runs without a working hypothesis against the share of slices seen; an answer given without any earlier hypothesis counts as the first one (15 runs). (b) Runs whose final answer was the first hypothesis. (c) Correct final answers after a wrong (n = 885) or right (n = 311) first hypothesis. (d) Free navigation (dark) vs navigation from the key slice (light; 1198 runs): share of slices seen at the first hypothesis, first hypothesis correct, first hypothesis kept, and correct final answers after a wrong or right first hypothesis; P values from generalized estimating equations clustered by case and adjusted for model; the difference in correct final answers after a wrong first hypothesis was absent within the model–case pairs with a wrong first hypothesis in both conditions (see Results). Numbers on the bars are percentages. Error bars, 95% CIs (bootstrap resampling cases).

Starting on the key slice brought the first hypothesis earlier (median, 3.2% of the slices [2.6%–3.7%]) and made it more often correct (37.5% [449/1198] vs 26.0%; OR, 1.73 [1.43–2.09]) and more often kept (73.0% [874/1198] vs 63.1%; OR, 1.59 [1.30–1.95]; both *P* < .001). Recovery after a wrong first hypothesis was rarer (6.3% [47/749] vs 12.4%; *P* < .001; Fig 4d) but not within the 392 model–case pairs with a wrong first hypothesis in both conditions (5.3% [38/714] vs 5.9% [43/725]; *P* = .64).

Keeping every image in context changed neither accuracy (28.9% [104/360] vs 30.0% [108/360]; *P* = .58) nor retention of the first hypothesis (64.4% [232/360] vs 66.1% [238/360]; *P* = .63; Table S6).

## Confidence

The models stated a median confidence of 90% (IQR, 82%–95%) in the 4778 final answers that stated one, whereas 38.9% of these answers (1861/4778) were correct, and correct and incorrect answers were given with similar confidence (median, 90% vs 88%; AUROC, 0.59 [0.55–0.63]). Calibration was poor: the Brier score was 0.453 (0.411–0.495), worse than the 0.238 of a constant confidence equal to the overall accuracy, and worse than that constant in every model and every condition. After the first hypothesis, mean confidence rose whether or not it was correct, and slightly more when it was wrong (Fig 5): from 35.6% to 85.1% after a wrong first hypothesis (885 runs) and from 42.3% to 89.3% after a right one (311 runs), a rise larger by 3.8 percentage points after a wrong hypothesis once adjusted for model (0.7–6.9; $P$ = .02). Confidence at the first hypothesis barely distinguished right from wrong hypotheses (AUROC, 0.60 [0.56–0.64]).

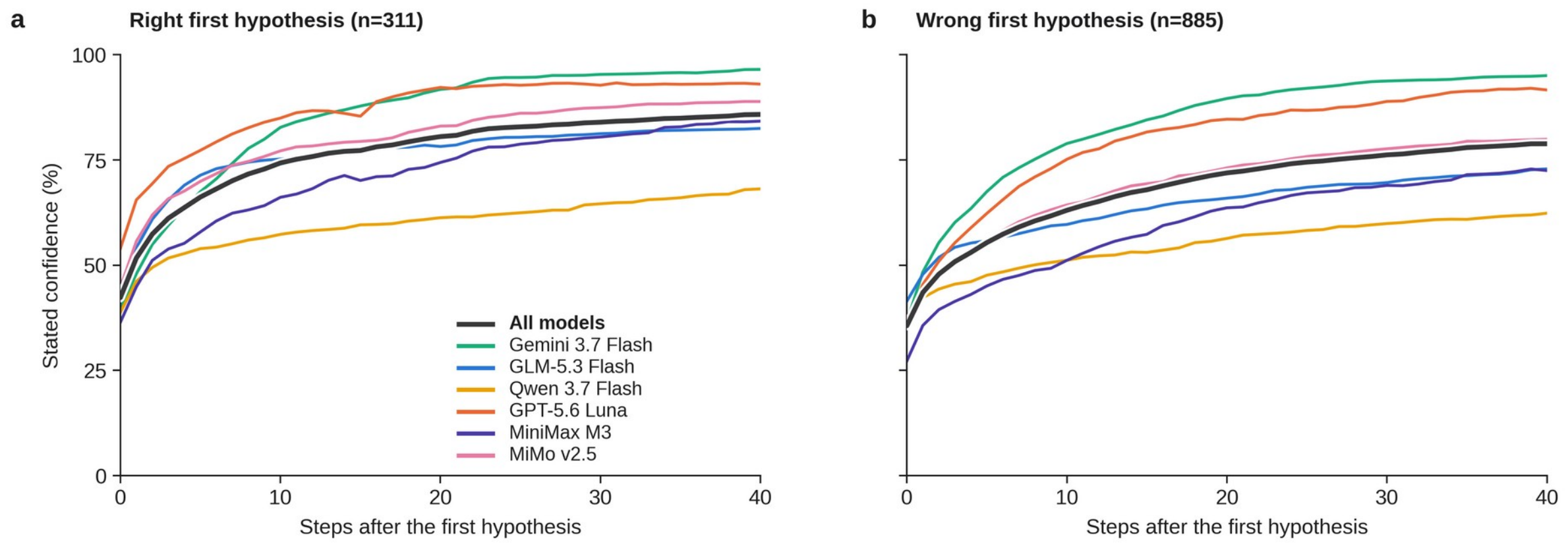

**Figure 5:** Stated confidence after the first working hypothesis (free navigation, 1196 runs). (a) Right first hypotheses (n = 311); (b) wrong first hypotheses (n = 885). Colored lines, mean of each model; dark grey, all models. Step 0 is the turn the hypothesis was stated; after the answer, the final confidence is carried forward.

## Discussion

In this study, VLM agents were about ten percentage points less accurate when navigating freely than with single requests, which in turn reached only about half the accuracy of radiologists on the same cases. This penalty parallels text-based clinical tasks, in which large language models diagnosed less accurately when they had to gather the history, laboratory results and imaging reports themselves (13) or elicit the history through conversation rather than read a vignette (20).

Across six models, a working hypothesis consistently emerged early in navigation and, in most runs, was not revised as new slices were encountered. The models other than Gemini 3.7 Flash went on to view most of the examination, but in a single uniform sweep, whereas radiologists spent more time on the key slices than on other slices. Search was therefore not the limiting step, which was more likely interpreting the key slices or revising incorrect hypotheses. This pattern resembles premature closure and anchoring in human readers (21) but probably arises differently: in humans, premature closure typically cuts the search short, whereas these models searched on and still held their first hypothesis, with confidence rising whether or not it was correct. Plausible causes are specific to language models: every turn is conditioned on the model's own earlier hypothesis, textual priors from the indication and anatomy can outweigh weak visual evidence (22,23), and training rewards confident answers over abstention (24). In physicians, too, errors arise during analytical as well as intuitive reasoning, and they are reduced by better knowledge rather than by awareness of bias (25). For these models, the missing knowledge may be the ability to read the images accurately.

In our no-image control, the models still reached about 40% of their accuracy with all slices. This mirrors the MIRAGE study (22), in which frontier VLMs described images they had never received and drew diagnoses from them, and other reports of a text-dominant reading of radiology cases by VLMs (3,23,26). Together, these results suggest that current agents draw a substantial part of their answers from the clinical context, anatomy and examination type, and that the images rarely lead them to the correct diagnosis once this first guess is wrong.

Two recent preprints that placed agents in real viewers point in the same direction. In ABRA (12), agents executed viewer commands almost perfectly but often failed to localize findings, and MedOpenClaw (11) showed that answer-only scores overestimate competence once the supporting evidence is checked. Our agents also navigated competently and displayed the key slices in most runs, yet most of their answers were wrong. About half of their wrong answers placed an abnormality in the correct organ, a share that fell when the images were withheld. This suggests that the images can lead the models to the right organ, but not yet to the right diagnosis.

VLM performance in medical imaging has usually been benchmarked on static, expert-curated image collections (1,2,27), where part of the score may not depend on the images at all (22,23). Our results show how fragile the scores of these models are: changing only the slice they started from changed their accuracy, while their median confidence stayed between 85% and 90%. As agentic AI systems move toward clinical radiology workflows (10), high performance on curated datasets cannot be assumed to translate into autonomous clinical competence (26–28), and confident outputs should not be read as reliability.

This study has limitations. First, the 100 studies came from an online teaching archive rather than clinical practice, and some were published before the knowledge cutoffs of the three models that report one, although accuracy did not differ between cases published before and after these cutoffs. Second, all tested models were competitive on benchmarks, but the thousands of runs required precluded testing their largest versions. Third, sending all slices in a single request removes navigation but also changes the input and removes search and stopping decisions. Starting from the key slice, which changes only the starting point, isolates part of the effect of navigation. Fourth, the agents had to state a working hypothesis at every turn, which stayed in their context and may itself have encouraged early commitment. Finally, only the primary diagnosis was graded, whereas radiology reports often give a differential diagnosis.

In conclusion, autonomous navigation of volumetric radiology studies exposes failure modes in VLM agents, notably early hypotheses that are usually kept and seldom corrected when wrong, that curated image sets do not reveal. Evaluating medical AI agents requires moving beyond static curated benchmarks toward dynamic environments that test active information gathering, reasoning iteration, and hypothesis revision.

## Data Availability

The code of the agent environment and of the analyses is available at https://github.com/bastienLeGuellec/radioagent. Run logs, grades and reader logs can be made available upon reasonable request to the corresponding author. All cases are public Radiopaedia cases (links in Table S1).

**Funding**

This study received no specific funding.

# Supplemental Material

Diagnostic Accuracy and Navigation Behavior of Vision-Language Model Agents Reading Emergency CT and MRI Examinations

## Appendix S1: Extended Methods

**Case selection.** Radiopaedia was searched by system (central nervous system, cardiac, chest, gastrointestinal, head and neck, musculoskeletal, obstetrics, paediatrics, spine, trauma, urogenital and vascular) and modality (CT and MRI). Series from another examination of the case (for example, an earlier CT) were added when the diagnosis required them. Stacks were shown with the window chosen by the case authors, which neither radiologists nor models could adjust. Table A1.1 describes the nine readers.

**Models.** The price limit kept the 4800 multi-step runs affordable. Published knowledge cutoffs were February 2026 for GPT-5.6 Luna, March 2026 for Gemini 3.7 Flash and May 2025 for MiMo v2.5; GLM-5.3 Flash, Qwen 3.7 Flash and MiniMax M3 publish none.

**Validation of the judge.** The 500 runs were sampled across the 24 model–condition combinations. Two radiologists (J.L. and B.L.G.) graded each answer independently, blinded to model, condition and judge verdict, agreed with each other on 83.2% of runs (Cohen κ = 0.69) and settled their 84 disagreements by consensus. Of four candidate judges, Claude Sonnet 5 agreed best with the consensus (agreement, 91.0%; κ = 0.83), as closely as each radiologist alone (91.8% and 91.0%; κ = 0.84 and 0.83).

**Behavioral measures.** Dwell ratio on key slices: share of navigation actions with a key slice on screen divided by the key slices' share of the examination. In the check of the segmentation of working hypotheses, the Methods give agreement on whether the first hypothesis was kept; on whether it was correct, agreement with the language model was κ = 0.79 and 0.65, and between the two radiologists κ = 0.67.

**Table A1.1: Radiologist readers.**

| Reader | Training level | Experience in radiology, y | Subspecialty | Correct answers (30 cases) |
|---|---|---|---|---|
| R1 | Senior resident (in training) | 3 | Musculoskeletal imaging | 22/30 (73.3%) |
| R2 | Senior resident (in training) | 4 | Chest imaging | 23/30 (76.7%) |
| R3 | Senior resident (in training) | 4 | Neuroradiology | 18/30 (60.0%) |
| R4 | Board-certified radiologist | 6 | Head and neck imaging | 22/30 (73.3%) |
| R5 | Board-certified radiologist | 7 | Neuroradiology | 23/30 (76.7%) |
| R6 | Board-certified radiologist | 7 | Neuroradiology | 24/30 (80.0%) |
| R7 | Board-certified radiologist | 8 | Abdominal imaging | 26/30 (86.7%) |
| R8 | Board-certified radiologist | 10 | Abdominal imaging | 24/30 (80.0%) |
| R9 | Board-certified radiologist | 10 | Pediatric imaging | 26/30 (86.7%) |

## Appendix S2: Agent Prompts, Tool Schemas and Run Parameters

This appendix gives the exact text sent to the models and to the grading model, the tool schemas, and the parameters of every run. Prompts are reproduced verbatim from the study code; text in angle brackets is filled in for each case. Every run file also stores its system prompt, parameters, each model reply and each feedback message (code repository: https://github.com/bastienLeGuellec/radioagent).

### S2.1. Run Parameters

**Table A2.1: Settings common to all runs.**

| Setting | Value |
|---|---|
| Access | OpenRouter API, one code path for all six models |
| Temperature | 0 |
| Reasoning | Provider default (no reasoning-effort parameter sent) |
| Images | JPEG, quality 85, longest side at most 512 pixels, sent as base64 data URLs |
| Images kept in context (navigation) | The 3 most recent slices; each earlier image is replaced by the text “[slice image shown earlier - no longer displayed]”. All text turns are kept. |
| Actions per turn | One; extra actions in the same reply were ignored |
| Step limit (navigation) | ceil(1.2 × total number of slices in the examination, all series) turns; then one forced request for the final answer (S2.2) |
| Images per single request | All slices up to 400 images (100 for Qwen 3.7 Flash); larger examinations were subsampled at a uniform stride (9 runs) |
| Invalid reply | One reprompt asking for a valid reply |
| Transient API errors | Up to 4 retries; every error logged in the run file |
| Runs without a usable answer | 18 of 4800 runs (provider content filter or no final answer), counted as incorrect |

**Table A2.2: Navigation runs (free navigation and navigation from the key slice) by model.**

| Model | Navigation runs | Turns, median (IQR) | Runs reaching the step limit | Runs with ≥1 invalid action | Runs without a usable answer (all conditions) |
|---|---|---|---|---|---|
| Gemini 3.7 Flash | 400 | 24 (14–34) | 0 | 3 | 0 |
| GPT-5.6 Luna | 400 | 38 (27–53) | 5 | 1 | 0 |
| GLM-5.3 Flash | 400 | 53 (38–75) | 41 | 1 | 0 |
| Qwen 3.7 Flash | 394 | 66 (41–91) | 77 | 129 | 16 |
| MiniMax M3 | 400 | 68 (43–95) | 151 | 91 | 0 |
| MiMo v2.5 | 400 | 36 (27–54) | 23 | 16 | 2 |

**Table A2.3: Navigation runs that reached the step limit (n = 297).**

| | **Answer forced at the step limit** | **Run continued to three times the number of slices** |
|---|---|---|
| Correct answers among these runs, % (n) | 28.3 (84 of 297) | 26.6 (79 of 297) |
| Accuracy, free navigation, % (n) | 32.6 (391 of 1200) | 32.5 (390 of 1200) |
| Accuracy, navigation from the key slice, % (n) | 37.9 (455 of 1200) | 37.6 (451 of 1200) |
| OR, navigation from the key slice vs free navigation (95% CI) | 1.27 (1.11–1.46) | 1.26 (1.10–1.43) |

Note.—Step limit, 1.2 times the number of slices. Runs that did not reach it were correct in 36.2% (762 of 2103; *P* = .08 vs runs at the limit). Forced vs continued answers, *P* = .36.

## S2.2. Navigation: System Prompt

The system prompt is sent with every request. The acquisition list gives each series name and its number of slices; for example, for case 5:

```
  - axial_t1 (16 slices)
  - axial_t2_fs (18 slices)
  - coronal_t1 (15 slices)
  - coronal_t2_fs (15 slices)
```

```
You are a radiologist examining a medical imaging study by navigating through its
slices.

Clinical indication: <clinical indication>

Available acquisitions (use these EXACT names with SWITCH):
  - <acquisition name> (<number of slices> slices)
  - ...

Each turn you receive the current slice image and must reply with ONLY a JSON object:
{"check": "...", "adjust": "...", "plan": "...", "working_hypothesis": "current best
diagnostic hypothesis, or 'none yet'", "confidence": 0-100, "command": "...", "final":
null or {"abnormal": true/false, "primary": "main diagnosis or 'Normal study'",
"differential": ["...", "..."], "confidence": 0-100}}

Field meaning:
- check: describe what you see on the current slice and whether it matches your
expectation.
- adjust: update your understanding of the case (anatomical landmarks, abnormalities,
and their slice numbers).
- plan: your current objective, the information still missing, and your next action.
- working_hypothesis: your single current best diagnosis given everything seen so far
("none yet" if you have no hypothesis).
- confidence: how confident you are in that hypothesis, 0-100.
- command: ONE of: UP (previous slice); DOWN (next slice); SWITCH <acquisition name>
(change series); DIAGNOSIS (finish the examination; also fill the "final" field).
- final: null, EXCEPT when command is DIAGNOSIS — then fill it: {"abnormal":
true/false, "primary": "main diagnosis or 'Normal study'", "differential": ["...",
"..."], "confidence": 0-100}

Guidance:
- First survey the study, then return to suspicious regions, and cross-reference other
acquisitions when useful.
- A completely normal study is a valid and important conclusion: if the study is
normal, finish with "abnormal": false and "primary": "Normal study".
- In "differential", list plausible alternatives in decreasing order of likelihood (may
be empty if certain).
- Only issue DIAGNOSIS when further navigation would not change your conclusion.
```

User messages. Free navigation starts with: “The examination begins. No acquisition is selected yet: your first command must be SWITCH. No image is shown yet.” Navigation from the key slice starts with “The examination starts at a pre-selected position.” followed by the position and the slice image, and “You may navigate freely from here.” After each command, the next message gives the result of the command and the new position, with the slice image, for example:

```
Switched to 'axial_t1'. Now at slice 1/16. Acquisition 'axial_t1', slice 1/16.
```

Other feedback messages: “Moved to slice <k>/<n>.”, “Already at first slice (1/<n>).”, “Already at last slice (<n>/<n>).”, “Invalid acquisition ‘<name>’. Available: <names>”, “No acquisition selected. Use SWITCH first.” and, for an unrecognized command, “Unrecognized command ‘<command>’. Valid: UP, DOWN, SWITCH <name>, DIAGNOSIS.” Switching series opens it at slice 1. At the step limit, the agent receives:

```
The examination step limit is reached. You must now conclude. Respond with ONLY this
JSON object: {"abnormal": true/false, "primary": "main diagnosis or 'Normal study'",
```

```
"differential": ["...", "..."], "confidence": 0-100}
```

### S2.3. Navigation: System Prompt and Tool Definitions

```
You are a radiologist examining a medical imaging study by navigating through its
slices.

Clinical indication: <clinical indication>

Available acquisitions (use these EXACT names with switch_acquisition):
  - <acquisition name> (<number of slices> slices)
  - ...

Each turn you receive the current slice image and MUST call exactly one tool:
- move(direction): go to the previous ("up") or next ("down") slice.
- switch_acquisition(acquisition): change series.
- finalize_diagnosis(...): finish the examination with your conclusion.

Every move/switch call also requires your reasoning fields (check, adjust, plan,
working_hypothesis, confidence) — fill them honestly each turn.

Guidance:
- First survey the study, then return to suspicious regions, and cross-reference other
acquisitions when useful.
- A completely normal study is a valid and important conclusion (abnormal=false,
primary="Normal study").
- In "differential", list plausible alternatives in decreasing order of likelihood (may
be empty if certain).
- Only call finalize_diagnosis when further navigation would not change your
conclusion.
```

Tool schemas (the reasoning fields check, adjust, plan, working_hypothesis and confidence are required in every move and switch_acquisition call):

```
[
  {
    "type": "function",
    "function": {
      "name": "move",
      "description": "Move one slice up (previous) or down (next) in the current
acquisition.",
      "parameters": {
        "type": "object",
        "properties": {
          "check": {
            "type": "string",
            "description": "What you see on the current slice and whether it matches your
expectation."
          },
          "adjust": {
            "type": "string",
            "description": "Updated understanding of the case (landmarks, abnormalities,
slice numbers)."
          },
          "plan": {
            "type": "string",
            "description": "Current objective, missing information, and why you take this
action."
          },
          "working_hypothesis": {
            "type": "string",
            "description": "Single current best diagnosis, or 'none yet'."
          },
          "confidence": {
            "type": "integer",
            "minimum": 0,
            "maximum": 100,
            "description": "Confidence in the working hypothesis, 0-100."
          },
          "direction": {
            "type": "string",
            "enum": [
              "up",
              "down"
            ]
          }
        },
        "required": [
```

```
          "check",
          "adjust",
          "plan",
          "working_hypothesis",
          "confidence",
          "direction"
        ]
      }
    }
  },
  {
    "type": "function",
    "function": {
      "name": "switch_acquisition",
      "description": "Change to another acquisition/series (starts at its first slice, or
where you left it).",
      "parameters": {
        "type": "object",
        "properties": {
          "check": {
            "type": "string",
            "description": "What you see on the current slice and whether it matches your
expectation."
          },
          "adjust": {
            "type": "string",
            "description": "Updated understanding of the case (landmarks, abnormalities,
slice numbers)."
          },
          "plan": {
            "type": "string",
            "description": "Current objective, missing information, and why you take this
action."
          },
          "working_hypothesis": {
            "type": "string",
            "description": "Single current best diagnosis, or 'none yet'."
          },
          "confidence": {
            "type": "integer",
            "minimum": 0,
            "maximum": 100,
            "description": "Confidence in the working hypothesis, 0-100."
          },
          "acquisition": {
            "type": "string",
            "description": "Exact acquisition name."
          }
        },
        "required": [
          "check",
          "adjust",
          "plan",
          "working_hypothesis",
          "confidence",
          "acquisition"
        ]
      }
    }
  },
  {
    "type": "function",
    "function": {
      "name": "finalize_diagnosis",
      "description": "Finish the examination and give the final conclusion. A completely
normal study is a valid conclusion (abnormal=false, primary='Normal study').",
      "parameters": {
        "type": "object",
        "properties": {
          "abnormal": {
            "type": "boolean"
          },
          "primary": {
            "type": "string",
            "description": "Main diagnosis, or 'Normal study'."
          },
          "differential": {
            "type": "array",
            "items": {
              "type": "string"
            },
```

```
            "description": "Plausible alternatives, most likely first."
          },
          "confidence": {
            "type": "integer",
            "minimum": 0,
            "maximum": 100
          }
        },
        "required": [
          "abnormal",
          "primary",
          "differential",
          "confidence"
        ]
      }
    }
  }
]
```

User messages are the same as in S2.2, with “your first action must be switch_acquisition”. A reply without a tool call is counted as an invalid action and answered with “You must call exactly one tool (move, switch_acquisition, or finalize_diagnosis).” At the step limit, the agent receives “The examination step limit is reached. You must now conclude by calling finalize_diagnosis.” and only the finalize_diagnosis tool is offered. Whatever the tool description says, switching always opened a series at slice 1.

### S2.4. Single Request (All Slices or Key Slices)

System prompt:

```
You are a radiologist interpreting a medical imaging study.

Clinical indication: <clinical indication>

You are shown <n> slices from the study. Each image is preceded by a label giving its
acquisition name and slice number.

Reply with ONLY a JSON object:
{"abnormal": true/false, "primary": "main diagnosis or 'Normal study'", "differential":
["...", "..."], "confidence": 0-100}

Guidance:
- A completely normal study is a valid and important conclusion: if the study is
normal, answer "abnormal": false and "primary": "Normal study".
- In "differential", list plausible alternatives in decreasing order of likelihood (may
be empty if certain).
- confidence: how confident you are in the primary diagnosis, 0-100.
```

With tool calls, the schema line reads “(use the finalize_diagnosis tool)” and the finalize_diagnosis tool of S2.3 is required. The user message is “Study slices follow (<n> images).”, then each image preceded by its label (“<acquisition name> - slice <k>/<number of slices>”), in series order, and ends with “End of study. Give your conclusion as the JSON object.” (or “End of study. Call finalize_diagnosis with your conclusion.”).

### S2.5. Grading of Final Answers

Model: Claude Sonnet 5 (anthropic/claude-sonnet-5), temperature 0, one request per run, without the model name or reading condition.
System prompt:

```
You are an expert radiologist adjudicating an AI system's diagnosis against the true
diagnosis of a case.

You will be given:
- The true diagnosis (ground truth).
- The AI system's final answer: whether it called the study abnormal, its primary
diagnosis, and its differential list.

Judge whether the AI's PRIMARY diagnosis is clinically correct using this practical
test: would treating physicians reach the same management from the AI's diagnosis as
from the true diagnosis, because the two identify essentially the same underlying
pathophysiological process? Judge by that test, not by lexical similarity.

- "correct": the primary diagnosis is the same disease process as the true diagnosis,
even if it differs in severity, subtype, or an added non-essential detail, because that
difference would not change the immediate management. Example: "acute pancreatitis" is
CORRECT against a true diagnosis of "acute necrotizing pancreatitis" (same process,
severity subtype). The same applies to exact laterality, exact vessel/segment, or an
added incidental/secondary finding that doesn't change the core process.
- "incorrect": the primary diagnosis names a clinically distinct process, even one in
the same organ or region or with overlapping presentation, if it does not share the
same underlying pathophysiology and would not lead to the same management. Example:
"appendicitis" is INCORRECT against a true diagnosis of "diverticulitis" (different
```

```
process despite similar presentation). "Cholecystitis" is INCORRECT against a true
diagnosis of "cholelithiasis" alone, and vice versa (inflammation vs. stones alone --
different management).
- If the true diagnosis is a normal/no-abnormality finding, the AI is correct only if
it also concluded normal/no significant abnormality.
- Do not give credit merely because the correct diagnosis appears somewhere in the
differential list -- only the PRIMARY diagnosis is judged, unless it is a close
paraphrase.

Use "partial" SPARINGLY -- only when the primary diagnosis is genuinely impossible to
call correct or incorrect by the test above (e.g. it names the right general process
but is so vague or hedged that management would plausibly diverge either way). Do not
use "partial" as a default for "close but not exact" -- resolve that to correct or
incorrect using the same-pathophysiology/same-management test above.

Respond with ONLY a JSON object:
{"verdict": "correct" | "partial" | "incorrect", "rationale": "one sentence"}
```

User message:

```
True diagnosis: {ground_truth}

AI system's answer:
- Abnormal study: {abnormal}
- Primary diagnosis: {primary}
- Differential: {differential}

Judge the primary diagnosis against the true diagnosis.
```

A verdict of partial was counted as incorrect.

### S2.6. Segmentation of Working Hypotheses

Model: Claude Sonnet 5, temperature 0, one request per navigation run with at least two turns. The user message lists the working hypothesis at each step (consecutive identical hypotheses collapsed into a range of steps) and ends with the final answer, for example:

```
Steps (step or range of steps: working hypothesis):
1-25: none yet
26: Cholelithiasis, possibly symptomatic (biliary colic); cholecystitis not yet
confirmed.
27-28: Cholelithiasis, clinically possible biliary colic; cholecystitis not confirmed.
29: Cholelithiasis with possible early cholecystitis; biliary colic vs. cholecystitis.
30-42: Acute cholecystitis secondary to cholelithiasis (stone at gallbladder neck).
43-65: Acute cholecystitis secondary to cholelithiasis (impacted stone at gallbladder
neck).
FINAL ANSWER: Acute cholecystitis secondary to cholelithiasis (gallstone impacted at
the gallbladder neck)
```

(GLM-5.3 Flash, case 5, free navigation, run 1.) At a turn where the agent answered, its answer stands for the working hypothesis; “(given at the step limit)” is added to the last line when the answer was forced. System prompt:

```
You are analysing the step-by-step notes of an AI agent reading a CT or MRI
examination. At each step it recorded a working hypothesis (its current best diagnosis)
or said it had none yet; consecutive steps with identical notes are shown once, as a
range of steps. The last line is the diagnosis it finally answered.

Split the steps into contiguous segments during which the agent held the same
underlying diagnosis.
- Same diagnosis: synonyms, rewordings, added detail or specificity (stroke ->
ischaemic stroke -> left MCA infarction), or a vague lead or broad category the agent
then names (possible mass -> meningioma; knee injury -> patellar dislocation).
- New segment: only when the agent moves to a clinically different diagnosis (ischaemic
stroke -> intracranial haemorrhage).
- If a step lists several candidates, the first-listed one is the working diagnosis.
- "None yet", "pending", or a description of findings without a diagnosis -> "none
yet". A statement that the study is normal ("no acute abnormality", "normal study") is
the diagnosis "normal study".
- Give each segment a short label, and reuse exactly the same label whenever the agent
returns to a diagnosis it held before.

Then compare the final answer with the first segment that is not "none yet", using the
same rules.

Reply with only this JSON:
{"segments": [{"from_step": <int>, "to_step": <int>, "diagnosis": "<label or 'none
yet'>"}, ...],
 "final_answer_label": "<label for the final answer, reusing a segment label if it is
```

```
the same diagnosis>",
 "final_answer_same_as_first": <true/false, or null if no segment is a diagnosis>}
Segments must cover every step from 1 to the last, in order, with no gaps or overlaps.
```

A reply whose segments did not cover every step exactly once was returned to the model once with the problem stated; replies were otherwise used as given.

### S2.7. Grading of First Hypotheses

When the final answer was not the first hypothesis, the grading model of S2.5 graded the agent's own wording of its first hypothesis at the last step at which it was held (and, as a sensitivity analysis, at its first step), with the same system prompt. In the user message, the abnormal field reads "(not assessed)", the differential is empty, and this note is appended:

```
(The primary diagnosis above is the agent's working hypothesis at one step of its
reading, before it gave its answer; if it lists several candidates, judge the first-
listed one.)
```

### S2.8. Failure Analysis: Classification of Wrong Answers

Model: Claude Sonnet 5 (anthropic/claude-sonnet-5), temperature 0, text only, one request per answer graded not correct (3952 answers, with and without images), without the model name or reading condition. System prompt:

```
You are an expert radiologist analysing the errors of an AI system that read a CT or
MRI examination. Its primary diagnosis has already been judged NOT correct against the
true diagnosis. Classify the error; do not re-judge correctness.

1. Sites. "true_site": the organ or anatomical structure involved by the true
diagnosis. "ai_site": the organ or structure where the AI's primary diagnosis places
the abnormality ("none" if the AI found no abnormality). Use plain anatomical names
(e.g. "liver", "left kidney", "L4-L5 disc", "right middle cerebral artery territory").

2. "location", one of:
- "same_organ": the AI's abnormality is in the same organ or structure as the true
diagnosis, even if the disease is wrong (true: liver metastasis; AI: liver abscess).
- "different_organ": the AI's abnormality is in a different organ or structure (true:
liver metastasis; AI: splenic infarction). Neighbouring organs count as different
(gallbladder vs pancreas).
- "no_abnormality": the AI called the study normal or reported no significant
abnormality, while the true diagnosis is abnormal.
- "not_localized": the AI's diagnosis has no comparable site (e.g. "sepsis",
"metastatic disease", "trauma" without an organ).
- "overcall": the true diagnosis is a normal study with no named finding and the AI
reported an abnormality.
When the true diagnosis is a normal study that names a variant or incidental finding
(e.g. "Normal (calyceal diverticulum with calculi)", "Horseshoe kidney (normal CT)"),
treat that named finding as the true site and compare locations as for an abnormal
case; never use "overcall" for these.

3. Disease process of each diagnosis, "true_process" and "ai_process", one of:
"vascular" (ischaemia, infarction, haemorrhage, thrombosis, embolism, dissection,
aneurysm, torsion), "infectious_inflammatory", "neoplastic" (benign or malignant,
including cysts that are tumours), "traumatic" (fracture, contusion, laceration,
haemorrhage caused by trauma), "congenital_variant" (developmental anomalies, normal
variants, non-neoplastic cysts), "degenerative_mechanical" (degeneration, herniation,
obstruction, volvulus, hydrocephalus, foreign body, calculi), "other", or "none"
(normal study with no named finding / no abnormality).
Classify by the underlying cause, not the mechanism: torsion is "vascular"; a
spontaneous (non-traumatic) rupture or haemorrhage takes the category of its cause
(infection: "infectious_inflammatory"; tumour: "neoplastic"; no cause given:
"vascular"); "traumatic" only when trauma is the cause. A normal study that names a
variant takes the category of that variant (usually "congenital_variant").

Respond with ONLY a JSON object:
{"true_site": "...", "ai_site": "...", "location": "...", "true_process": "...",
"ai_process": "...", "rationale": "one sentence"}
```

User message:

```
True diagnosis: {ground_truth}
Radiopaedia case title: {title}
Examination: {modality}

AI system's answer (judged not correct):
- Abnormal study: {abnormal}
- Primary diagnosis: {primary}
```

## Appendix S3: Statistical Analysis

The unit of analysis was the run, and runs of the same case were treated as correlated. Runs without a usable answer (18 of 4800: provider content filter) were counted as incorrect. Accuracy was compared between conditions with logistic generalized estimating equations (GEE; exchangeable correlation, clustered by case, adjusted for model): an omnibus Wald test, then the six pairwise comparisons, with Holm-adjusted P values and odds ratios (ORs). The same model, with a condition-by-domain interaction, compared anatomical domains. Ninety-five percent CIs for proportions, medians and their differences came from bootstrap resampling of cases (10 000 resamples), because runs of the same case are not independent; for comparisons with the radiologists, the bootstrap resampled both cases and readers, because every reader read every case: each resample drew the 30 cases and, independently, the nine readers with replacement, and applied the same case draw to the radiologists' reads and to the model runs, so that the comparison stayed paired by case (the six models were not resampled). The two-sided P value was twice the share of resampled differences on the far side of zero. Agreement between the two runs of a model was measured with Fleiss κ, which treats the two runs as interchangeable, for each model and condition, and agreement among radiologists with Fleiss κ.

Exploration measures are reported as medians with IQRs and were compared between models with Gaussian GEE clustered by case. The association between a correct first hypothesis and a correct final answer was estimated with logistic GEE adjusted for model; a within-case OR was obtained by adding the case mean of the predictor as a covariate. Calibration of the final confidence was assessed with the Brier score (mean squared difference between stated confidence, scaled from 0 to 1, and correctness), compared with that of a constant forecast equal to the overall accuracy, and the area under the receiver operating characteristic curve (AUROC) of confidence for a correct answer. The rise in confidence after a right versus a wrong first hypothesis was compared with Gaussian GEE clustered by case and adjusted for model. The comparison of conditions was the primary analysis; other analyses were exploratory. A two-sided $P < .05$ indicated statistical significance. Analyses used Python 3.12 (statsmodels 0.14.5, SciPy 1.16.1).

## Table S1: Cases

| Case | Domain | Diagnosis | Clinical indication | Radiopaedia case (radiopaedia.org/cases/…) |
|---|---|---|---|---|
| 1 | Abdominal | Acute appendicitis | Abdominal pain. | acute-appendicitis-partially-obstructed-lumen-1 |
| 2 | Abdominal | Acute appendicitis (pregnancy) | Abdominal pain. | acute-appendicitis-in-pregnancy-9 |
| 3 | Abdominal | Peritonitis | Abdominal pain. | peritonitis-3 |
| 4 | Abdominal | Gastritis | Abdominal pain. | gastritis-3 |
| 5 | Abdominal | Acute cholecystitis | Abdominal pain. | acute-calculus-cholecystitis-2 |
| 6 | Abdominal | Perforated peptic ulcer | Abdominal pain. | perforated-peptic-ulcer-8 |
| 7 | Abdominal | Necrotizing pancreatitis | Abdominal pain. | necrotizing-pancreatitis-18 |
| 8 | Abdominal | Bowel obstruction | Abdominal pain. | closed-loop-obstruction-adhesive-bands |
| 9 | Abdominal | Acute diverticulitis | Abdominal pain. | acute-diverticulitis-26 |
| 10 | Neurologic | Aneurysmal subarachnoid haemorrhage | Headache. | subarachnoid-haemorrhage-aneurysmal |
| 11 | Neurologic | Post-traumatic intracranial haemorrhage | Road traffic accident. | intracranial-haemorrhage-6 |
| 12 | Neurologic | Acute ischaemic stroke | Acute neurological deficit. | hyperacute-stroke-with-hyperdense-mca-sign |
| 13 | Neurologic | Basilar artery occlusion | Altered consciousness. | basilar-artery-occlusion-9 |
| 14 | Neurologic | Maxillary fractures | Head trauma. | maxillary-fractures-1 |
| 15 | Neurologic | Normal (CT head) | Head trauma. | normal-ct-head-and-cervical-spine-pediatric-1 |
| 16 | Neurologic | Cortical vein thrombosis | Headache. | isolated-subacute-cortical-vein-thrombosis |
| 17 | Thoracic | Pulmonary embolism | Dyspnea. | acute-pulmonary-embolism-7 |
| 18 | Thoracic | Aortic dissection | Chest pain. | progression-of-a-stanford-b-intramural-hematoma-to-dissection |
| 19 | Thoracic | Bronchial foreign body | Acute respiratory distress. | left-main-bronchus-foreign-body-with-obstructive-pneumonia |
| 20 | Thoracic | Normal (chest CT) | Cough. | gown-tie-artifact |
| 21 | Thoracic | Pneumonia | Cough. | chlamydia-pneumonia-2 |
| 22 | Neurologic | C2 fracture | Trauma. | extension-teardrop-fracture-of-c2 |
| 23 | Neurologic | Lumbar disc herniation | Low back pain. | lumbar-disc-herniation-with-caudal-migration-3 |
| 24 | Musculoskeletal and trauma | Periprosthetic fracture | Mechanical fall. | periprosthetic-femoral-fracture-vancouver-type-b2-5 |
| 25 | Musculoskeletal and trauma | Spondylodiscitis | Low back pain. | spondylodiscitis-39 |
| 26 | Neurologic | Normal (incidental Tarlov cyst) | Low back pain. | tarlov-cyst-16 |
| 27 | Abdominal | Ovarian torsion | Abdominal pain. | ovarian-torsion-125 |
| 28 | Abdominal | Ureteric calculus | Abdominal pain. | ureteric-calculus-14 |
| 29 | Abdominal | Normal (calyceal diverticulum with calculi) | Abdominal pain. | calyceal-diverticulum-with-calculi |
| 30 | Musculoskeletal and trauma | Splenic trauma (grade V) | Abdominal trauma. | splenic-injury-grade-v |
| 31 | Neurologic | Pericallosal cerebral infarct | Left-sided hemiparesis. | pericallosal-cerebral-infarct |
| 32 | Neurologic | Quadrigeminal cistern arachnoid cyst | Headache after head trauma. | quadrigeminal-cistern-arachnoid-cyst-11 |
| 33 | Neurologic | Normal CT-scan (arachnoid granulation in right transverse sinus) | Headache. | normal-cerebral-ct-angiography |
| 34 | Neurologic | Ruptured Anterior Communicating Artery Aneurysm | Found comatose. | ruptured-anterior-communicating-artery-aneurysm-5 |
| 35 | Neurologic | Meningioma - intraosseous component | Enlarging forehead mass. | meningioma-intraosseous-component |
| 36 | Neurologic | Posterior reversible encephalopathy syndrome (PRES) | Postpartum vomiting, drowsiness and hypertension. | posterior-reversible-encephalopathy-syndrome-pres-36 |
| 37 | Neurologic | Herpes simplex encephalitis | Unsteady gait and dizziness. | herpes-simplex-encephalitis-31 |
| 38 | Neurologic | Basilar Artery Occlusion | Sudden headache and aphasia. | basilar-artery-occlusion-6 |
| 39 | Neurologic | High-velocity non-penetrating traumatic brain injury | Gunshot injury to the head. | high-velocity-non-penetrating-traumatic-brain-injury |
| 40 | Neurologic | Normal brain scan | Acute dizziness and headache. | normal-brain-ct-4 |
| 41 | Neurologic | Acute infarct - right putamen | Acute dysarthria. | acute-infarct-right-putamen |
| 42 | Neurologic | MCA hyperacute infarction | Sudden left-sided weakness and speech difficulty. | mca-hyperacute-infarction |
| 43 | Neurologic | Arachnoid cyst | Recurrent seizures and headache. | arachnoid-cyst-69 |
| 44 | Neurologic | MCA infarction with hyperdense MCA sign | Left-sided hemiplegia. | hyperdense-mca-sign-24 |
| 45 | Neurologic | Subarachnoid haemorrhage | Severe headache. | subarachnoid-hemorrhage-19 |
| 46 | Neurologic | Dural venous sinus thrombosis | Left-sided numbness, weakness and loss of consciousness. | dural-venous-sinus-thrombosis-33 |
| 47 | Neurologic | Cytotoxic lesions of the corpus callosum (CLOCCs) | Worsening chronic migraine. | cytotoxic-lesions-of-the-corpus-callosum-cloccs-10 |
| 48 | Neurologic | Tuberculum sellae meningioma | Bitemporal hemianopsia. | tuberculum-sellaemeningioma-3 |
| 49 | Neurologic | Artery of Percheron Infarction | Obtunded after cardiac surgery. | artery-of-percheron-infarction-14 |
| 50 | Neurologic | Normal Pressure Hydrocephalus | Confusion. | normal-pressure-hydrocephalus-37 |
| 51 | Thoracic | Pulmonary metastases from lung cancer | Persistent dry cough. | pulmonary-metastases-from-lung-adenocarcinoma |
| 52 | Thoracic | Cardiogenic pulmonary oedema | Chest pain and dyspnea. | cardiogenic-pulmonary-oedema-4 |
| 53 | Thoracic | Varicose Bronchiectasis | Chronic cough in a smoker. | varicose-bronchiectasis-3 |
| 54 | Thoracic | Varicose and cystic bronchiectasis | Productive cough and wheezing. | varicose-and-cystic-bronchiectasis-1 |
| 55 | Thoracic | Saddle pulmonary embolism | Dyspnea and tachycardia. | saddle-pulmonary-embolism-33 |
| 56 | Thoracic | Pneumocystis jirovecii pneumonia | Fever and cough in a transplant recipient. | pneumocystis-jirovecii-pneumonia-12 |
| 57 | Thoracic | Pulmonary fibrosis - UIP pattern | Persistent dyspnea and cough. | pulmonary-fibrosis-uip-pattern-5 |
| 58 | Thoracic | Foreign body aspiration | Sudden cough and dyspnea. | foreign-body-aspiration-13 |
| 59 | Thoracic | Lymphangioleiomyomatosis | Progressive exertional dyspnea. | lymphangioleiomyomatosis-38 |
| 60 | Thoracic | Pyopneumothorax | Fever, cough and chest pain. | pyopneumothorax-9 |
| 61 | Thoracic | Lung transplant - invasive aspergillosis | Fever and cough in a lung transplant recipient. | lung-transplant-invasive-aspergillosis |
| 62 | Thoracic | Pulmonary tuberculosis | Cough, fever and night sweats. | pulmonary-tuberculosis-91 |
| 63 | Thoracic | Left upper lobe atelectasis (small cell lung carcinoma with upper left lobe bronchus compression) | Shortness of breath. | left-upper-lobe-atelectasis |
| 64 | Thoracic | Usual Interstitial Pneumonia | Breathlessness. | usual-interstitial-pneumonia-30 |
| 65 | Thoracic | Large Pneumothorax | Chest pain and dyspnea after trauma. | large-pneumothorax-9 |
| 66 | Abdominal | Closed-loop small bowel obstruction | Acute abdominal pain. | closed-loop-small-bowel-obstruction-adhesive |
| 67 | Abdominal | Omental infarction | Left iliac fossa pain. | omental-infarction-40 |
| 68 | Abdominal | Emphysematous pyelonephritis | Reduced consciousness and raised inflammatory markers. | emphysematous-pyelonephritis-60 |
| 69 | Abdominal | Appendicitis + abscess | Fever and right lower quadrant pain. | ruptured-acute-appendicitis |
| 70 | Abdominal | Adrenal hematoma | Fall onto right chest. | traumatic-adrenal-haematoma |
| 71 | Abdominal | Acute pancreatitis | Epigastric pain. | acute-pancreatitis-uncinate-process |
| 72 | Abdominal | Splenic rupture | Fever, night sweats and weight loss. | spontaneous-splenic-rupture-due-to-cmv-infection |
| 73 | Abdominal | Fournier's gangrene | Sepsis and deteriorating condition. | fourniers-gangrene-10 |
| 74 | Abdominal | Epiploic appendagitis | Right lower abdominal pain. | epiploic-appendagitis-132 |
| 75 | Abdominal | Colocolic intussusception secondary to lipoma | Left flank pain and hematochezia. | colocolic-intussusception-secondary-to-lipoma |
| 76 | Abdominal | Incarcerated femoral hernia | Right lower quadrant pain. | incarcerated-femoral-hernia-4 |
| 77 | Abdominal | Acute necrotizing pancreatitis | Severe abdominal pain and raised amylase. | acute-necrotizing-pancreatitis-6 |
| 78 | Abdominal | Ureterolithiasis | Abdominal pain and hematuria. | ureterolithiasis-24 |
| 79 | Abdominal | Horseshoe kidney (normal CT) | Incidental finding on CT angiography. | horseshoe-kidney-60 |
| 80 | Abdominal | High-grade blunt hepatic injury (AAST grade V) | Blunt abdominal trauma with hypotension. | high-grade-blunt-hepatic-injury-aast-grade-v |
| 81 | Abdominal | Sigmoid volvulus | Acute abdominal pain. | sigmoid-colon-volvulus |
| 82 | Abdominal | Acute bowel ischemia | Abdominal pain and raised lactate. | acute-bowel-ischemia-2 |
| 83 | Abdominal | Midgut volvulus | Abdominal pain and distension. | midgut-volvulus-27 |
| 84 | Abdominal | Iatrogenic bladder injury following laparoscopic hysterectomy | Pelvic pain after hysterectomy. | iatrogenic-bladder-injury-following-laparoscopic-hysterectomy |
| 85 | Musculoskeletal and trauma | Spinal epidural hematoma | Severe sciatica. | spinal-epidural-hematoma-6 |
| 86 | Musculoskeletal and trauma | Lumbar disc extrusion with caudal migration | Low back pain radiating to the left. | lumbar-disc-extrusion-with-caudal-migration |
| 87 | Musculoskeletal and trauma | Sacral insufficiency fractures | Pelvic pain. | vitamin-d-deficiency-osteomalacia-with-multiple-insufficiency-fractures |
| 88 | Musculoskeletal and trauma | Spondylodiscitis - thoracic | Back pain and fever. | spondylodiscitis-thoracic-1 |
| 89 | Musculoskeletal and trauma | Proximal hamstring tear | Sudden hamstring pain during sport. | proximal-hamstring-tear |
| 90 | Musculoskeletal and trauma | Distal femoral osteomyelitis with subperiosteal abscess | Fever and thigh pain. | distal-femoral-osteomyelitis-with-subperiosteal-abscess |
| 91 | Musculoskeletal and trauma | Spinal epidural abscess | Sepsis and back pain. | spinal-epidural-abscess-8 |
| 92 | Musculoskeletal and trauma | Posterior hip dislocation | Road traffic accident with hip deformity. | posterior-hip-dislocation-15 |
| 93 | Musculoskeletal and trauma | Achilles tendon rupture | Popping sensation and ankle pain during sport. | achilles-tendon-tear-38 |
| 94 | Musculoskeletal and trauma | Complete distal quadriceps tendon rupture with partial MPFL injury | Knee trauma with inability to extend. | complete-distal-quadriceps-tendon-rupture-with-partial-mpfl-injury |
| 95 | Musculoskeletal and trauma | Acetabular fractures | Fall from height. | pelvic-fractures-bilateral-acetabular |
| 96 | Musculoskeletal and trauma | Patellar tendon rupture | Sports injury. | patellar-tendon-rupture-and-repair |
| 97 | Musculoskeletal and trauma | Flexion tear drop fracture and unilateral facet dislocation | Trauma with inability to walk. | flexion-tear-drop-fracture-and-unilateral-facet-dislocation |
| 98 | Musculoskeletal and trauma | Posterior shoulder dislocation | Shoulder pain. | posterior-shoulder-dislocation-41 |
| 99 | Musculoskeletal and trauma | Cervical DISH with fracture | Neck pain after a fall. | cervical-dish-with-fracture |
| 100 | Musculoskeletal and trauma | Patellar dislocation-relocation | Sports injury. | patellar-dislocation-relocation |

## Table S2: Diagnostic Accuracy by Anatomical Domain and Reading Condition

| Domain | Cases | Free navigation | Navigation from key slice | All slices, single request | Key slices, single request | All conditions | Radiologists* |
|---|---|---|---|---|---|---|---|
| Abdominal | 31 | 25.5 (16.7–35.8) | 30.1 (20.2–40.3) | 34.7 (23.4–46.2) | 35.5 (24.7–46.8) | 31.5 (21.9–41.5) | 75.9 (82 of 108) |
| Neurologic | 30 | 35.8 (26.1–46.1) | 40.8 (28.9–52.8) | 44.2 (33.1–55.3) | 46.4 (35.0–58.1) | 41.8 (31.7–52.4) | 80.0 (72 of 90) |
| Thoracic | 20 | 45.0 (28.7–60.8) | 54.2 (38.8–68.8) | 57.9 (44.2–71.2) | 55.4 (41.2–69.2) | 53.1 (39.0–67.2) | 77.8 (35 of 45) |
| Musculoskeletal and trauma | 19 | 25.9 (14.5–39.0) | 28.9 (18.4–40.8) | 34.6 (21.1–49.1) | 34.2 (21.0–48.7) | 30.9 (19.7–43.2) | 70.4 (19 of 27) |
| All domains | 100 | 32.6 (26.7–38.8) | 37.9 (31.8–44.4) | 42.2 (35.7–48.8) | 42.5 (36.2–48.9) | 38.8 (32.9–44.7) | 77.0 (208 of 270) |

Note.—Data are percentages of correct answers, with 95% CIs (bootstrap resampling cases) in parentheses; each model and condition contributes two runs per case. Domain effect, $P = .05$; condition-by-domain interaction, $P = .93$ (logistic generalized estimating equations clustered by case, adjusted for model and condition). *The 30 cases read by the radiologists, graded by the model judge; numbers in parentheses are correct reads of all reads.

## Table S3: Agreement between the Two Runs of Each Model, by Reading Condition

| Model | Free navigation | Navigation from key slice | All slices, single request | Key slices, single request |
|---|---|---|---|---|
| Gemini 3.7 Flash | 0.63 (0.47, 0.78) 82% | 0.82 (0.70, 0.92) 91% | 0.70 (0.55, 0.84) 85% | 0.70 (0.55, 0.83) 85% |
| GLM-5.3 Flash | 0.68 (0.53, 0.82) 85% | 0.68 (0.52, 0.82) 84% | 0.74 (0.60, 0.86) 87% | 0.58 (0.42, 0.74) 79% |
| Qwen 3.7 Flash | 0.62 (0.45, 0.77) 83% | 0.75 (0.60, 0.87) 88% | 0.76 (0.61, 0.88) 89% | 0.76 (0.62, 0.88) 89% |
| GPT-5.6 Luna | 0.55 (0.36, 0.72) 81% | 0.75 (0.58, 0.88) 90% | 0.69 (0.54, 0.83) 85% | 0.85 (0.74, 0.94) 93% |
| MiniMax M3 | 0.53 (0.32, 0.72) 84% | 0.48 (0.29, 0.66) 77% | 0.73 (0.58, 0.85) 87% | 0.79 (0.66, 0.90) 90% |
| MiMo v2.5 | 0.32 (0.11, 0.52) 74% | 0.48 (0.27, 0.65) 77% | 0.48 (0.29, 0.65) 77% | 0.47 (0.28, 0.64) 76% |

Note.—Data are Fleiss κ, which treats the two runs as interchangeable, (95% CI) for correct versus incorrect answers between the two runs of each model on the 100 cases, with the percentage of cases in which both runs agreed below. CIs are from a bootstrap that resampled cases.

## Table S4: Diagnostic Accuracy without Images (No-Image Control, All 100 Cases)

| Measure | Free navigation | All slices, single request | Slice labels, no images | Indication only |
|---|---|---|---|---|
| **All models pooled** | | | | |
| Accuracy, all cases, % | 32.6 (391 of 1200) [26.7, | 42.2 (506 of 1200) [35.7, | 17.2 (103 of 600) [12.0, | 11.3 (68 of |
| | 38.8] | 48.8] | 22.5] | 600) [7.2, 15.7] |
| Accuracy, abnormal reference, % | 30.7 (343 of 1116) | 41.3 (461 of 1116) | 13.4 (75 of 558) | 8.4 (47 of 558) |
| Accuracy, 54 cases unsolved without images, % | 18.2 (118 of 648) | 27.8 (180 of 648) | 0 | 0 |
| Accuracy, other 46 cases, % | 49.5 (273 of 552) | 59.1 (326 of 552) | 37.3 (103 of 276) | 24.6 (68 of 276) |
| OR vs all slices, single request (95% CI) | — | Reference | 0.28 (0.20–0.38) | 0.17 (0.11–0.25) |
| OR vs free navigation (95% CI) | Reference | — | 0.42 (0.31–0.58) | 0.26 (0.18–0.38) |
| No diagnosis, images reported missing, % | — | — | 5.5 (33 of 600) | 28.2 (169 of 600) |
| Confidence, median | 90 | 88 | 70 | 50 |
| **By model: accuracy, %** | | | | |
| Gemini 3.7 Flash | 44.0 | 52.5 | 27.0 | 22.0 |
| GLM-5.3 Flash | 38.5 | 52.5 | 13.0 | 3.0 |
| Qwen 3.7 Flash | 34.5 | 34.5 | 16.0 | 17.0 |
| GPT-5.6 Luna | 30.5 | 41.5 | 9.0 | 6.0 |
| MiniMax M3 | 22.0 | 38.5 | 22.0 | 15.0 |
| MiMo v2.5 | 26.0 | 33.5 | 16.0 | 5.0 |

Note.—Six models. Free navigation and all slices in a single request: main study, two runs per case (1200 runs per condition). No-image controls: one run per case (600 runs per condition), October 2026, same models, settings, prompts and judge. Slice labels, no images: the all-slices request with every series name and slice label but no image; the model was not told that the images were missing. Indication only: the clinical indication alone. ORs from logistic GEE clustered by case and adjusted for model; $P < .001$ for all four comparisons. 54 cases unsolved without images: cases that no model answered correctly in either no-image condition. Data in parentheses are numbers of runs; for accuracy on all cases, 95% CIs (bootstrap resampling of cases) are in brackets.

## Table S5: Outcome of Every Run by Reading Condition

| Outcome | Free navigation | Navigation from key slice | All slices, single request | Key slices, single request | Slice labels, no images | Indication only |
|---|---|---|---|---|---|---|
| Correct | 32.6 (391) | 37.9 (455) | 42.2 (506) | 42.5 (510) | 17.2 (103) | 11.3 (68) |
| Wrong, same organ | 24.6 (295) | 24.9 (299) | 21.6 (259) | 25.7 (308) | 12.2 (73) | 6.8 (41) |
| Wrong, different organ | 23.1 (277) | 21.4 (257) | 18.4 (221) | 18.0 (216) | 19.3 (116) | 9.3 (56) |
| Wrong, no comparable site | 0.4 (5) | 0.2 (3) | 0.2 (3) | 0.5 (6) | 0.5 (3) | 0.7 (4) |
| Abnormal case called normal | 18.5 (222) | 14.4 (173) | 16.4 (197) | 13.1 (157) | 45.2 (271) | 43.7 (262) |
| Abnormality reported on a normal case | 0.5 (6) | 0.7 (8) | 0.7 (8) | 0.0 (0) | 0.2 (1) | 0.0 (0) |
| No answer | 0.3 (4) | 0.4 (5) | 0.5 (6) | 0.2 (3) | 0.0 (0) | 0.0 (0) |
| No diagnosis, images reported missing | 0.0 (0) | 0.0 (0) | 0.0 (0) | 0.0 (0) | 5.5 (33) | 28.2 (169) |
| Same organ, % of wrong answers naming an abnormality | 51.6 (295 of 572) | 53.8 (299 of 556) | 54.0 (259 of 480) | 58.8 (308 of 524) | 38.6 (73 of 189) | 42.3 (41 of 97) |
| Same disease category, % of the same answers | 47.7 (273 of 572) | 46.9 (261 of 556) | 53.8 (258 of 480) | 52.7 (276 of 524) | 56.6 (107 of 189) | 67.0 (65 of 97) |

Note.—Data are percentages of runs, with numbers of runs in parentheses (1200 runs per condition with images, two per case; 600 per no-image condition, one per case); each run has exactly one outcome. No-image answers that gave no diagnosis because the images were missing were not classified and are shown in their own row. Wrong answers were classified by the study judge from the reference and the final diagnosis. Last two rows: among wrong answers that placed an abnormality in an organ (same or different). Disease categories: vascular, infectious or inflammatory, neoplastic, traumatic, congenital or variant, degenerative or mechanical, other. Same organ, free navigation vs all slices in a single request: OR, 0.87 (95% CI: 0.68, 1.11); without vs with images: OR, 0.52 (95% CI: 0.38, 0.72) (logistic generalized estimating equations clustered by case, adjusted for model).

## Table S6: Sensitivity Analysis with Every Image Kept in Context (Free Navigation, Cases Read by the Radiologists)

| Measure | Three most recent images | All images kept | OR (95% CI) | P |
|---|---|---|---|---|
| **All models pooled** | | | | |
| Accuracy, % | 30.0 (108 of 360) | 28.9 (104 of 360) | 0.95 (0.78–1.15) | .58 |
| First hypothesis kept as final answer, % | 66.1 (238 of 360) | 64.4 (232 of 360) | 0.93 (0.68–1.26) | .63 |
| First hypothesis correct, % | 24.7 (89 of 360) | 24.2 (87 of 360) | 0.97 (0.79–1.19) | .77 |
| Correct final answer after a wrong first hypothesis, % | 10.3 (28 of 271) | 8.8 (24 of 273) | 0.82 (0.59–1.14) | .23 |
| Slices viewed, median % (IQR) | 96.5 (69.6–100.0) | 92.9 (62.3–100.0) | — | .44 |
| Actions per slice viewed, median (IQR) | 1.00 (1.00–1.06) | 1.00 (1.00–1.01) | — | .31 |
| Dwell ratio on key slices, median (IQR) | 1.00 (0.96–1.09) | 1.00 (1.00–1.10) | — | .25 |
| Turns, median (IQR) | 42 (26–65) | 40 (28–61) | — | .22 |
| **By model: accuracy / first hypothesis kept, %** | | | | |
| Gemini 3.7 Flash | 35.0 / 66.7 | 35.0 / 75.0 | — | — |
| GLM-5.3 Flash | 40.0 / 71.7 | 36.7 / 65.0 | — | — |
| Qwen 3.7 Flash | 28.3 / 73.3 | 28.3 / 78.3 | — | — |
| GPT-5.6 Luna | 30.0 / 55.0 | 26.7 / 60.0 | — | — |
| MiniMax M3 | 18.3 / 58.3 | 18.3 / 66.7 | — | — |
| MiMo v2.5 | 28.3 / 71.7 | 28.3 / 41.7 | — | — |

Note.—Six models, two runs per case (360 runs per arm); the original runs kept the three most recent images in context, the repeat every image seen so far (up to each provider's image limit), with a step limit of three times the number of slices. ORs and P values from logistic (binary measures) or Gaussian (continuous

measures) generalized estimating equations clustered by case and adjusted for model; per-model values are descriptive. Runs without a final answer were counted as incorrect. Dwell ratio, share of turns with a key slice on screen divided by the key slices' share of the examination (1 for a uniform sweep).

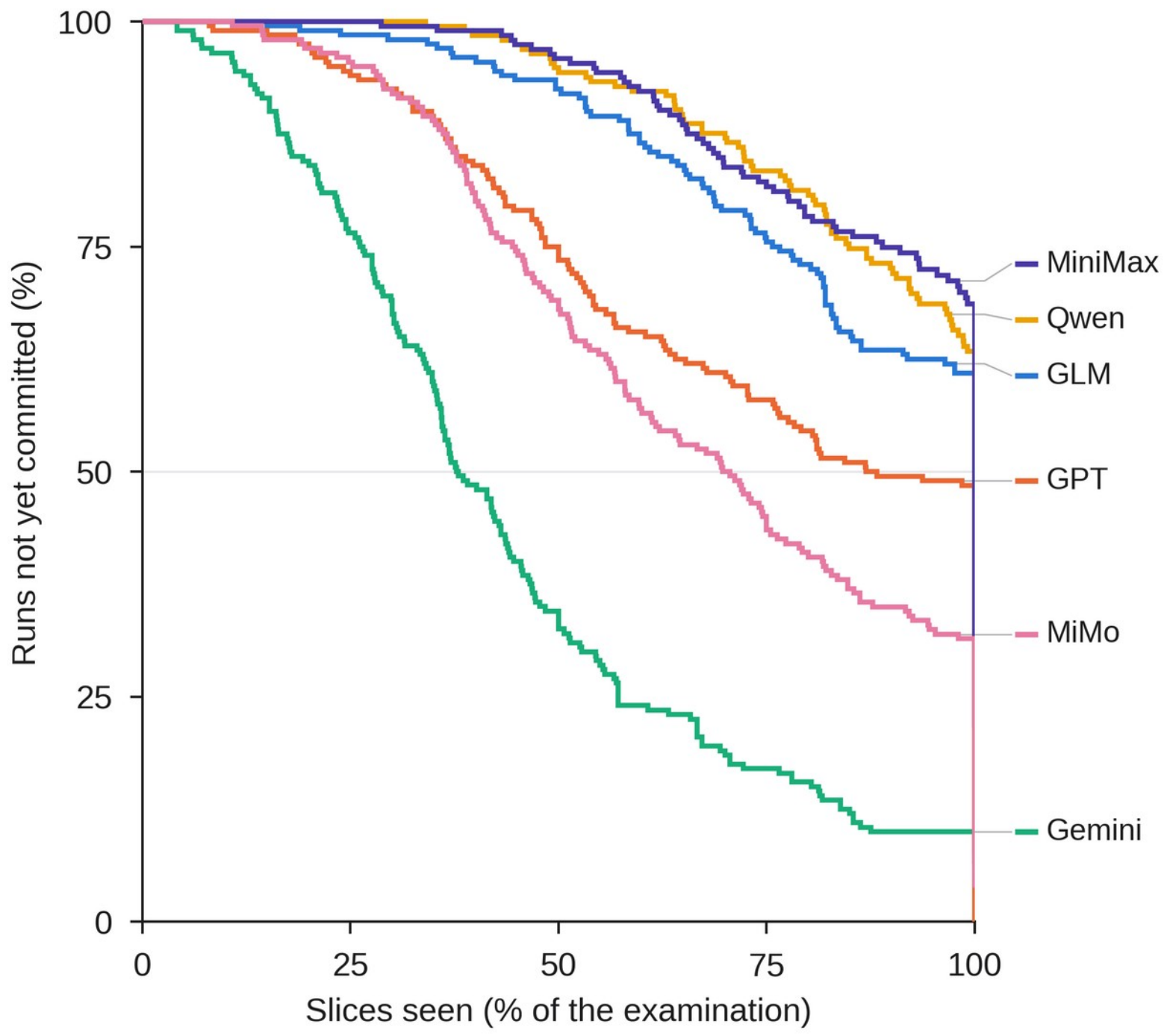

**Figure S1:** Share of the examination seen before answering, all 100 cases (free navigation; six models, 196–200 runs each). Kaplan-Meier curves as in Figure 3a.